\documentclass[letterpaper]{article} 
\usepackage{aaai2026}  
\usepackage{times}  
\usepackage{helvet}  
\usepackage{courier}  
\usepackage[hyphens]{url}  
\usepackage{graphicx} 
\usepackage{natbib}  
\usepackage{caption} 
\usepackage{algorithm}
\usepackage{algorithmic}

\usepackage{amsmath} 
\usepackage{amsfonts}
\usepackage{subcaption} 
\usepackage{graphicx}
\usepackage{multirow}
\usepackage{graphicx}
\usepackage{array}

\usepackage{makecell} 
\usepackage{adjustbox}
\usepackage{color, xcolor}
\usepackage{colortbl}

\usepackage{bbding}
\usepackage{pifont}
\usepackage{booktabs} 

\usepackage{newfloat}
\usepackage{listings}
\DeclareCaptionStyle{ruled}{labelfont=normalfont,labelsep=colon,strut=off} 
\floatstyle{ruled}
\newfloat{listing}{tb}{lst}{}
\floatname{listing}{Listing}
\title{Separation and Collaboration: Two-Level Routing Grouped Mixture-of-Experts for Multi-Domain Continual Learning}
\author{
    Jialu Zhou\textsuperscript{\rm 1},
    Dianxi Shi\textsuperscript{\rm 2}\thanks{Corresponding author.},
    Shaowu Yang\textsuperscript{\rm 1},
    Xinyu Wei\textsuperscript{\rm 1},
    Mingyue Yang\textsuperscript{\rm 1},
    Leqian Li\textsuperscript{\rm 1},\\
    Mengzhu Wang\textsuperscript{\rm 4},
    Chunping Qiu\textsuperscript{\rm 3}
}
\affiliations{
    \textsuperscript{\rm 1}College of Computer Science and Technology, National University of Defense Technology, Changsha, China\\
    \textsuperscript{\rm 2}Advanced Institute of Big Data, Beijing, China\\
	\textsuperscript{\rm 3}Intelligent Game and Decision Lab, Beijing, China\\
    \textsuperscript{\rm 4}Hebei University of Technology, Tianjin, China\\
    
     \{zhoujialu23, dxshi, shaowu.yang, xinyuwei, yangmingyue216, llq\}@nudt.edu.cn,\\
     202005223@stu.ncwu.edu.cn, chunping.qiu@aliyun.com
}

\usepackage{bibentry}

\begin{document}

\maketitle

\begin{abstract}
Multi-Domain Continual Learning (MDCL) acquires knowledge from sequential tasks with shifting class sets and distribution. Despite the Parameter-Efficient Fine-Tuning (PEFT) methods can adapt for this dual heterogeneity, they still suffer from catastrophic forgetting and forward forgetting. To address these challenges, we propose a Two-Level Routing Grouped Mixture-of-Experts (TRGE) method. Firstly, TRGE dynamically expands the pre-trained CLIP model, assigning specific expert group for each task to mitigate catastrophic forgetting. With the number of experts continually grows in this process, TRGE maintains the static experts count within the group and introduces the intra-group router to alleviate routing overfitting caused by the increasing routing complexity. Meanwhile, we design an inter-group routing policy based on task identifiers and task prototype distance, which dynamically selects relevant expert groups and combines their outputs to enhance inter-task collaboration. Secondly, to get the correct task identifiers, we leverage Multimodal Large Language Models (MLLMs) which own powerful multimodal comprehension capabilities to generate semantic task descriptions and recognize the correct task identifier. Finally, to mitigate forward forgetting, we dynamically fuse outputs for unseen samples from the frozen CLIP model and TRGE adapter based on training progress, leveraging both pre-trained and learned knowledge. Through extensive experiments across various settings, our method outperforms other advanced methods with fewer trainable parameters.
\end{abstract}


\section{Introduction}

In the dynamic open-world, machine learning needs to adapt to the continuously  changes. Traditional deep learning approaches typically rely on static datasets\cite{wang2023m, li2024instance, chen2024rsmamba}, which requires costly full retraining when new tasks appear. Continual Learning (CL) addresses this by incrementally updating models only with new data, but suffering from catastrophic forgetting of learned knowledge\cite{goodfellow2013empirical}. The more realistic Multi-Domain Continual Learning (MDCL), including Multi-Domain Task Incremental Learning (MTIL)\cite{yu2024select} and Multi-Domain Class Incremental Learning (MCIL)\cite{yu2024select}, further complicates this challenge with both class sets and domains varying across tasks. However, traditional CL methods rely on fixed task-specific classification heads\cite{tang2024mind} which is difficult to adapt for this dual shifts in classes and domains.

\begin{figure}[t]
\centering
\includegraphics[width=1\columnwidth]{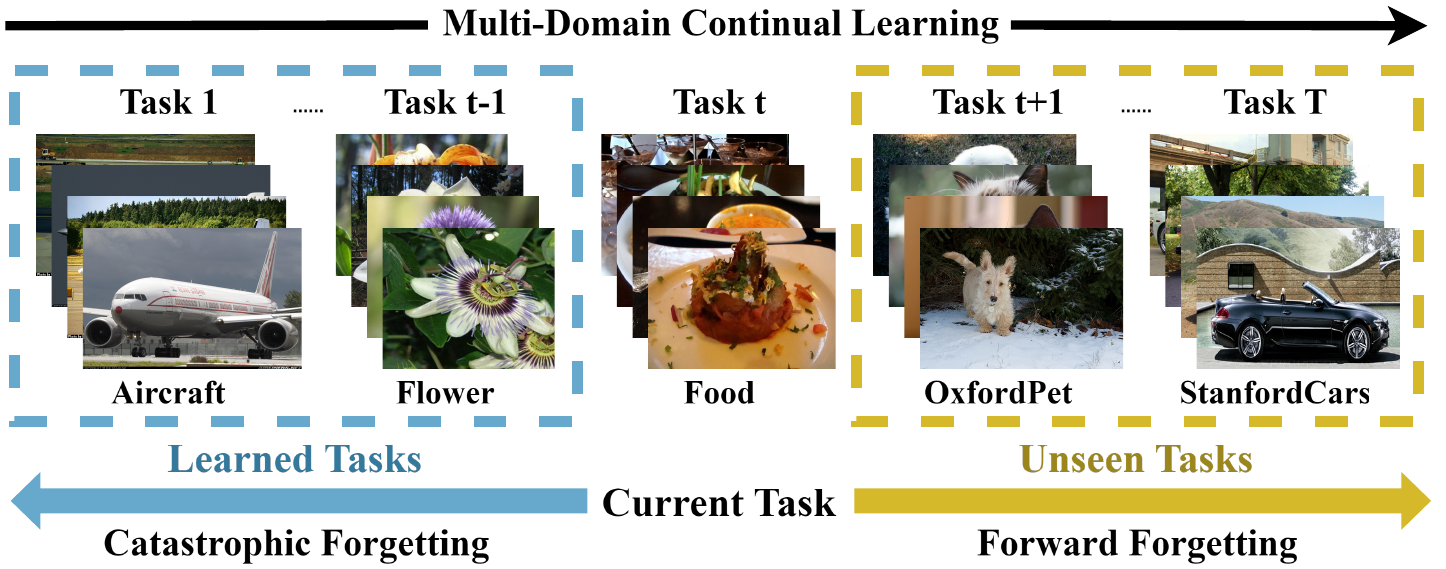}
\caption{Illustration of catastrophic forgetting and forward forgetting. Catastrophic forgetting loses knowledge of learned tasks (task $1$ to task $t-1$). Forward forgetting loses pretrained knowledge, degrading performance on unseen tasks (task $t+1$ to task $T$).}
\label{MDIL}
\end{figure}

Recently, pre-trained Vision-Language Models (VLMs) with remarkable cross-domain generalization and multi-modal alignment capabilities, such as CLIP\cite{radford2021learning}, shows significant potential to address MDCL problems. However, full-parameter fine-tuning of VLMs risks forward forgetting\cite{zheng2023preventing} which means the degradation of model's zero-shot ability, as shown in Figure \ref{MDIL}. To mitigate this, Parameter-Efficient Fine-Tuning (PEFT) methods\cite{gao2024clip,zaken2021bitfit} freeze pre-trained weights and quickly adapt for the downstream tasks by training a small number of parameters. But among existing methods, it’s hard to ensure both knowledge accumulation and transfer at the same time. For example, approaches based on parameter isolation\cite{wang2022s,chitale2023task} can reduce forgetting but will limit cross-task knowledge transfer. Alternatively, methods with shared prompt pool\cite{wang2022learning,park2024pre} promote the knowledge reuse, but as the prompts are trained with all tasks, it still risks forgetting. Based on this, we aim to develop a PEFT strategy that balances task-specific knowledge retention and cross-task knowledge transfer to mitigate catastrophic forgetting.

For this challenge, Mixture-of-Experts (MoE)\cite{jacobs1991adaptive,shazeer2017outrageously}, as a highly flexible and scalable architecture, shows great potential\cite{huai2025cl,rypesc2024divide}. With mutually independent experts and sparse gating, MoE can preserve task-specific knowledge in experts and achieve cross-task knowledge transfer by routing. However, the problems arises when applying the MoE with one router to CL. The static number of experts limits its knowledge capacity, conflicting with the incremental requirements of CL. Alternatively, dynamic expert expansion is prone to performance degradation due to routing policy overfitting and representational collapse\cite{gao2024higher}. Therefore, how to mitigate the conflict between MoE's inherent structure and the dynamic demands of incremental knowledge learning needs to be resolved.

To overcome these challenges, we propose a Two-Level Routing Grouped Mixture-of-Experts (TRGE) method. Firstly, we assign task-specific experts for each task by expanding the pre-trained model and freeze them in subsequent learning to protect learned knowledge. As the total number of experts grows with tasks coming, we group the experts for the same task together in a fixed size and introduce an intra-group router within each group to maintain the complexity of intra-group routing, avoiding the routing policy overfitting. Furthermore, we propose an inter-group routing policy. The inter-group router selects the main and assistant groups respectively based on task identifier and the task prototype distances, then combines their outputs with scaled weights to achieve the task collaboration. Secondly, as the main group selection relies on the correct task identifiers of samples, we propose a Semantic-based Task Recognition (STR) method for it. STR leverages Multimodal Large Language Models (MLLMs) with its powerful abilities in multimodal understanding and alignment to generate task descriptions from class sets and recognize the identifies with input image and all tasks’ descriptions. This method avoids mistakes from ambiguous feature boundaries in existing feature-based methods, significantly improving the accuracy. Thirdly, we dynamically combine the outputs of the pre-trained model and TRGE adapter based on training phase to classify unseen samples with both pre-trained and task-specific knowledge, mitigating forward forgetting.

Our contributions can be summarized into the following three aspects:
\begin{itemize}
\item We propose a Two-Level Routing Grouped MoE adapter. Through expert grouping and two-level routing, it ensures task collaboration and mitigate catastrophic forgetting at the same time.
\item  We introduce a Semantics-based Task Recognition method. Leveraging MLLMs to accurately recognize task identifiers based on class sets and input samples which can improve the rationality of inter-group routing and correctly identify the unseen samples.
\item  We design a dynamic fusion mechanism which adaptively weights outputs from the pre-trained CLIP and TRGE adapter, effectively integrating knowledge from the pre-training and task learning stages to alleviate the forward forgetting.
\end{itemize}

\section{Related Work}
\subsection{Continual Learning}
Continual learning research primarily addresses three settings: Task-Incremental Learning (TIL), Class-Incremental Learning (CIL), and Domain-Incremental Learning (DIL)\cite{yang2025recent}. Existing methods can be categorized into regularization, replay, structure expansion and PEFT. Regularization-based approaches constrain parameter updates via penalty terms or knowledge distillation\cite{kirkpatrick2017overcoming,benzing2022unifying, yu2024select} for mitigating forgetting, but suffering from high computational complexity and sensitivity to task sequence. Replay-based methods retain or generate historical samples to recall knowledge from previous tasks\cite{kumari2022retrospective, park2021class, zhuang2022multi} but are limited by data privacy and buffer capacity. Structure expansion methods dynamically grow networks with task-specific modules to increase the knowledge capacity\cite{xue2022meta, douillard2022dytox} but inherently struggles to balance parameter efficiency against cross-task knowledge transfer. PEFT strategies have recently gained traction for CL\cite{smith2023coda, wang2022learning, park2024pre} which update small number of parameters on frozen pre-trained models. However, they also struggle to balance forgetting resistance and knowledge transfer in multi-domain incremental scenarios. Thus, we propose an incremental framework based on MoE, which isolates expert groups to prevent catastrophic forgetting and enables flexible cross-task collaboration via an inter-group router, achieving a better performance trade-off.

\subsection{Mixture of Experts}
MoE combines the outputs of different specialized experts through a sparse gating network\cite{shazeer2017outrageously}. This design naturally aligns with CL objectives: expert independence mitigates catastrophic forgetting while sparse activation enables dynamic knowledge composition. Therefore, recent works have adapted MoE for continual learning. U-TELL\cite{solomon2024u} dynamically instantiates task expert modules and employs a task assigner for expert selection. MoE-Adapter\cite{yu2024boosting} expands CLIP with task-specific router, using MoE routing to selectively activate expert modules for each tasks. However, existing research has paid little attention to the negative impact of routing scale expansion on model performance during continual learning\cite{gao2024higher}. Our method fixes the intra-group routing complexity within fixed-size expert groups to mitigate the impact of incremental scenarios on routing effectiveness and employs lightweight inter-group router to maintain the flexibility of MoE.

\begin{figure*}[t]
\centering
\includegraphics[width=1\textwidth]{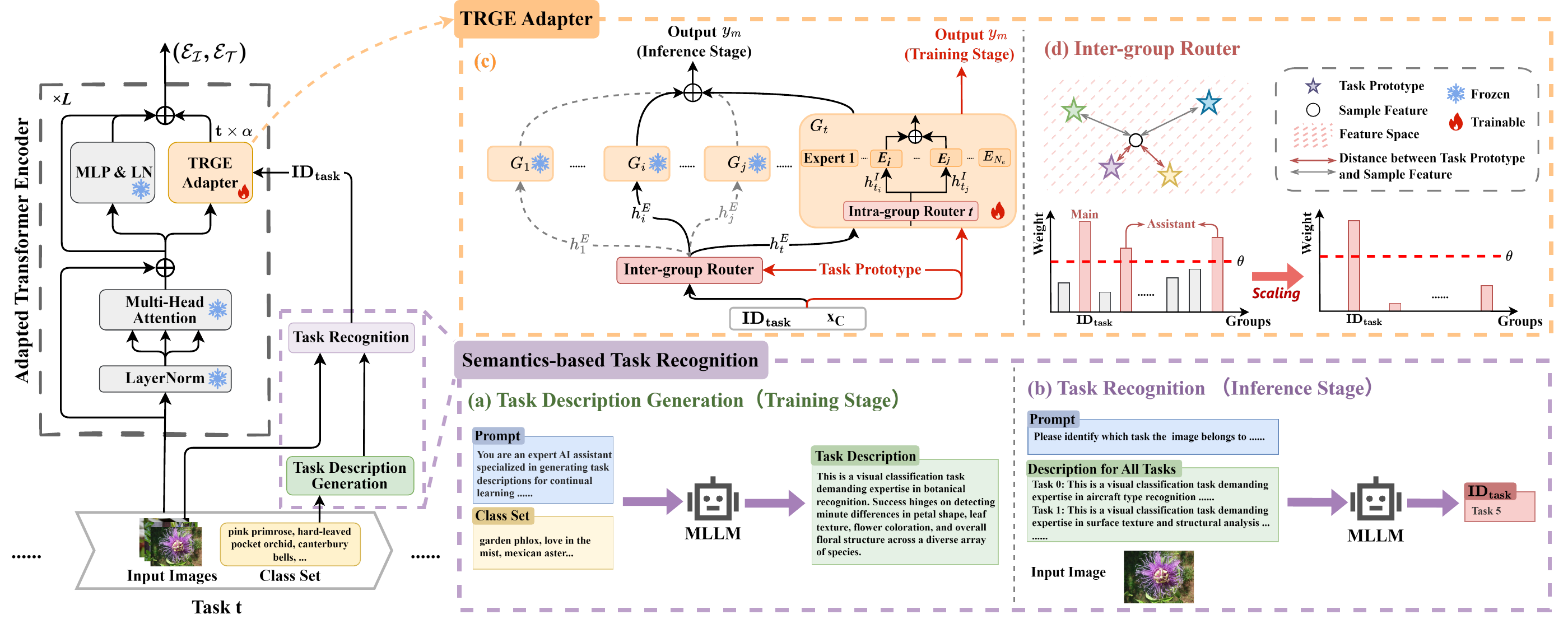}
\caption{The overall framework of the our method. Semantic-Based Task recognition includes two aspects: generating task descriptions during training (a) and task recognition during inference (b). The TRGE Adapter leverages expert grouping and a two-level routing strategy to save task knowledge and fine-tune the pre-trained model (c). It shows inter-group routing based on task identifiers and task prototype distances with a weight scaling strategy (d). Finally, combining the output of the pre-trained module and the TRGE adapter based on dynamic fusion weights: $t \times \alpha$.}
\label{framework}
\end{figure*}

\section{Method}
\subsection{Task Definition}
Continual learning sequentially acquires knowledge from an ordered sequence of $T$ tasks ${\left \{ {\mathcal{T}_t} \right \} ^T_{i=1}}$. Each task $\mathcal{T}_t=\left \{ D_t, C_t\right \} $ consists of a class set $C_t$ and a dataset $D_t=\left \{ x^t_i,y^t_i \right \} ^ {N_t} _{i=1}$, where $x^t_i$ donating input sample, $ y^t_i \in C_t $ is the corresponding label and ${N_t}$ is the total number of samples per task. On both MTIL and MCIL, tasks exhibit distinct domains($O_i\neq O_j, i \neq j$). Differently, MCIL requires disjoint class sets ($ C_i \cap C_j = \emptyset, i \neq j $), while tasks may share overlapping classes ($C_i \cap C_j \neq \emptyset, i \neq j$) on MTIL. Meanwhile, task identifier $ ID_{task}$ is only known on MTIL which requires the model to predict across all learned classes $ {\textstyle \bigcup_{i=1}^{t}C_i}$ on MCIL during the inference.

\subsection{Framework Overview}
To address the challenges of catastrophic forgetting and forward forgetting in MDCL, we propose a Two-level Routing Grouped MoE method upon the CLIP which contains parallel encoders for images and text. Following CLIP, we assign input to the class with highest probability computed by the cosine similarity between the image embedding $\mathcal{E_I}$ for input images and text embedding $\mathcal{E_T}$ for class sets.

The overall framework of our method is shown in Figure \ref{framework}. During the training phase, the task description of the current task is generated by MLLM with the prompt and class set (Figure \ref{framework}(a)). Then, TRGE adapter freezes the previous groups and trains a new group for the current task to mitigate catastrophic forgetting. Meanwhile, the mean of all sample feature from the current task is stored as the task prototype in the inter-group router (Figure \ref{framework}(c)). During the inference phase, MLLM recognizes the task identifier (Figure \ref{framework}(b)) through its powerful multimodal understanding on task descriptions and sample image, avoiding mistakes caused by statistical features information. Then, the inter-group router confirms the main group based on the task identity, while the assistant groups are selected by the distance between the sample features and stored task prototypes. Groups’ outputs are combined based on scaled weights (Figure \ref{framework}(d)) which achieve effective collaboration between tasks. In addition, if the sample is unseen, the outputs of the pre-trained model and the TRGE adapter are adaptively fused based on the training phase, utilizing the pre-trained and task knowledge for forward forgetting alleviation.

\subsection{Two-Level Routing Grouped Mixture-of-Experts}
The TRGE adapter mainly consists of two parts: an inter-group router and the task-specific expert groups. Each expert group only learns from one task which prevent forgetting in continual learning. Meanwhile, experts grouping ensures the intra-group router connecting to a fixed number of experts, avoiding the risks of increasing routing complexity. Further, dynamically selecting and combining the relevant groups through inter-group router achieves the effective task collaboration.

\subsubsection{Task-Specific Expert Groups.}For each new task, a new group is initialized for adaptation. Considering the number of experts growing with the tasks continual training, we employ parameter-efficient LoRA\cite{hu2022lora} as the expert network to mitigate the parameter overhead, following \cite{yu2024boosting}.

The task-specific expert group $G_t$ contains an intra-group router $R^I_t$ and a set of expert networks $\left \{ E^t_i \right \} ^{N_e}_{i=1}$ with the same structure, where $N_e$ is the total number of experts within a group. $R^I_t$ calculates the gating weights $H^I_t =\{h^I_{t_i}\}_{i=1}^{N_e}$ for experts through a gating function:
\begin{equation}  
H^I_t=Softmax(Topk(x_{C}\cdot W^t))
\end{equation}
where $ W_t\in \mathbb{R}^{d \times N_e}$ is a learnable weight matrix. To improve efficiency, we refer to\cite{wang2022learning}, using the $[CLS]$ token $x_C \in \mathbb{R}^{d}$ as the input. $Topk(\cdot )$ selects the $k$ most relevant experts. $Softmax(\cdot)$ normalizes the weights of the selected experts. With $H^I_t$, outputs from expert groups $ E^t_i(\cdot) $ can be weighted combined:
\begin{equation}  
G_t(x_C)=\sum_{i=1}^{N_e} h^I_{t_i} E^t_i(x_C)
\end{equation}
where $G_t(\cdot)$ denotes the output of group $G_t$.

To prevent the task knowledge from being corrupted by subsequent task training, only the group corresponded to current task will be trained while previous groups are frozen, alleviating forgetting. Simultaneously, each intra-group router operates on the fixed experts count, remaining unaffected by the dynamically expanding architectural so that the grouped MoE can be more applicable for continual learning.

\subsubsection{Dynamical Inter-group Routing.}With each group only learns from one task, beneficial knowledge transfer across tasks are limited. To address this, we propose an inter-group routing mechanism which dynamically selects and weights outputs from both the main group which contains the corresponding task knowledge to input and highly relevant assistant groups, enabling cross-task knowledge collaboration.

As groups adding for new tasks, the complexity of inter-group routing grows accordingly. To minimize the computational burden, the routing should be lightweight. Therefore, we implement a weighting method base on the distance between input and task prototypes that independently evaluates groups' relevance. Only the assistant groups with higher relevance than threshold $\theta$ or the main group corresponded by task identifier $ID_{task}$ are activated. Specifically, task prototype $P_t$ is the mean feature of all inputs within task $t$.
\begin{equation}  
P_t=\frac{1}{N_t}\sum_{x\in D_t}x_C
\end{equation}

Concatenate $P_t$ to the task prototype matrix $W^g \in \mathbb{R}^{t \times d}$, where the $t$-th row corresponds to the prototype $P_t$. Relevance is measured by the Euclidean distance between input $x_C$ and task prototypes:
\begin{equation}
d_i = \left \|x_C-P_i\right \| _2
\end{equation}
\begin{equation}
v_i=d_{max}-d_i
\end{equation}
where $d_{max} = \max(d_1, d_2, \ldots, d_t)$ is the maximum distance. The group weight ${H^E}' = \text{Softmax}(V)$ are obtained by normalizing relevance set $V = \{v_i\}_{i=1}^t$. However, the normalization will reduce the distinction between groups, leading to reduced routing rationality. Especially when excessive weights are assigned to low-relevance groups, it may introduce noise interference. Thus, we introduce a weight scaling mechanism. We first set a threshold $\theta$ to select highly relevant assistant groups and then scale weights: the weight of main group ($i = ID_{task}$) is set to 1, selected assistant group weights $ {h^E_t}' - \theta$ and set others to be $-\infty$.
\begin{equation}  
{h^E_i}'= \begin{cases} 1 & \text{if  }  i = ID_{task} \\{h^E_i}' -\theta & \text{if  } i \neq ID_{task} \text{ and } {h^E_i}' > \theta \\-\infty & \text{if  } i \neq ID_{task} \text{ and } {h^E_i}' \leq \theta\end{cases}
\end{equation} 
 After scaling, normalizing weights $ H^E= \text{Softmax}({H^E}') $ to ensure the sum of all weights is 1. The output of TRGE adapter $ y_m$ is the weighted combination of expert groups.
\begin{equation}  
y_m=\sum_{i=1}^{t} h^E_i G_i (x_C)
\end{equation}

\begin{figure}[t]
\centering
\includegraphics[width=0.9\columnwidth]{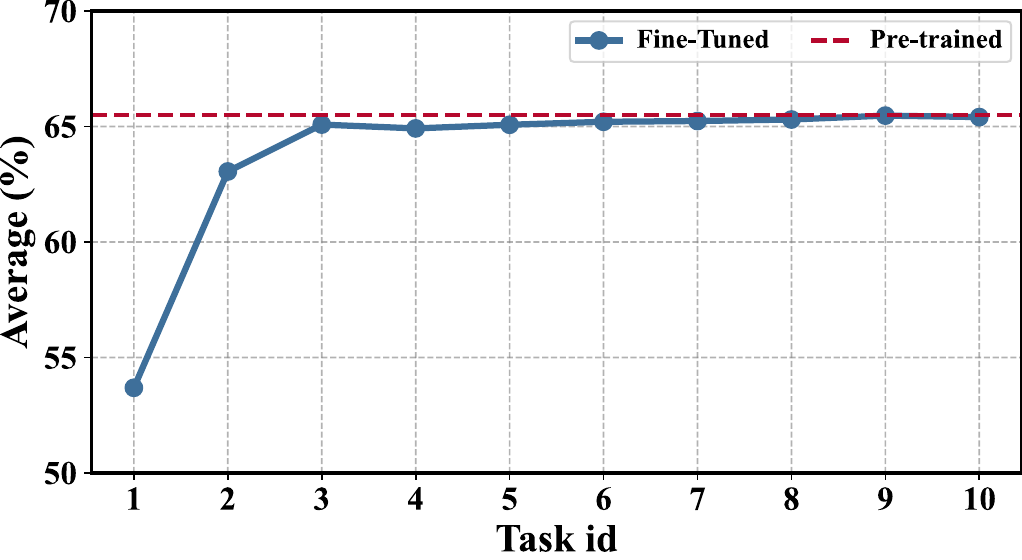}
\caption{Zero-shot prediction accuracy of fine-tuned and pretrained models on SUN397 dataset during different training phases.}
\label{zero-shot}
\end{figure}

\subsection{Semantics-based Task Recognition}
During inter-group routing, the activation of the main group depends on the accurate recognition for task identity of input which will affect the overall performance of TRGE adapter. However, traditional approaches (e.g. clustering\cite{huai2025cl}, feature reconstruction\cite{yu2024boosting} and so on) mainly based on feature statistics information which are prone to misidentification due to ambiguous feature distribution boundaries across tasks\cite{metzner2022classification}. To address this challenge, we propose a Semantics-based Task Recognition (STR) method. By leveraging MLLMs' semantic understanding and image parsing capabilities, STR achieves more precise task recognition with semantic information derived from class sets instead of statistic features.

To enrich task description information and enhance semantic distinguishability across different tasks, we utilize MLLMs to parse and integrate the discrete class set $C_t$ into a refined task description $s_t$.
\begin{equation}  
s_t=MLLM(C_t)
\end{equation}

\begin{table*}[t]
\centering
\small
\setlength{\tabcolsep}{5pt} 
\begin{tabular}{cl|c|ccccccccccc|c}
\hline
\multicolumn{2}{c|}{\textbf{Method}}                                       & \textbf{\# Param.}  & $\mathbf{D_1}$ &  $\mathbf{D_2}$ & $\mathbf{D_3}$ & $\mathbf{D_4}$ & $\mathbf{D_5}$ & $\mathbf{D_6}$ & $\mathbf{D_7}$ & $\mathbf{D_8}$ & $\mathbf{D_9}$ & $\mathbf{D_{10}}$ & $\mathbf{D_{11}}$ & \textbf{Mean}                        \\ \hline
\multicolumn{1}{c|}{}                           & Zero-shot                    &                                & 24.3                                  & 88.4                                  & 68.2                                  & 44.6                                  & 54.9                                  & 71.0                                  & 88.5                                  & 59.4                               & 89.0                                  & 64.7                                  & 65.2                               & 65.3                                        \\
\multicolumn{1}{c|}{\multirow{-2}{*}{CLIP}}     & Full Fine-tune               &                                & 62.0                                  & 95.1                                  & 89.6                                  & 79.5                                  & 98.9                                  & 97.5                                  & 92.7                                  & 99.6                               & 94.7                                  & 89.6                                  & 81.8                               & 89.2                                        \\ \hline
\multicolumn{1}{c|}{Qwen-VL}                    & In-Context                   &                                & 45.2                                  & 93.4                                  & 80.0                                  & 66.8                                  & 86.2                                  & 95.7                                  & 84.7                                  & 78.0                               & 79.2                                  & 85.6                                  & 65.6                               & 78.2                                        \\ \hline  \hline
\multicolumn{1}{c|}{}                           & Continual-FT                 &                                &                                       & 67.1                                  & 46.0                                  & 32.1                                  & 35.6                                  & 35.0                                  & 57.7                                  & 44.1                               & 60.8                                  & 20.5                                  & 46.6                               & 44.6                                                               \\
\multicolumn{1}{c|}{}                           & LwF                          & 211M                           &                                       & 74.5                                  & 56.9                                  & 39.1                                  & \underline{51.1}                            & 52.6                                  & 72.8                                  & 60.6                               & 75.1                                  & 30.3                                  & 55.9                               & 58.9                                                               \\
\multicolumn{1}{c|}{}                           & iCaRL                        & 211M                           &                                       & 56.6                                  & 44.6                                  & 32.7                                  & 39.3                                  & 46.6                                  & 68.0                                  & 46.0                               & 77.4                                  & 31.9                                  & 60.5                               & 50.4                                                               \\
\multicolumn{1}{c|}{}                           & ZSCL                         & 211M                           &                                       & 86.0                                  & 67.4                                  & \textbf{45.4}                         & 50.4                                  & 69.1                                  & 87.6                                  & \textbf{61.8}                      & 86.8                                  & 60.1                                  & \textbf{66.8}                      & 68.1                                                               \\
\multicolumn{1}{c|}{}                           & MoE-Adapter                  & 59.8M                          &                                       &   \underline{87.9}                            & \underline{68.2}                            & 44.4                                  & 49.9                                  &   \underline{70.7}                            & \textbf{88.7}                         & 59.7                               & \textbf{89.1}                         & \textbf{64.5}                         & 65.5                               &   \underline{68.9}                                                         \\  
\rowcolor{gray!30} \multicolumn{1}{c|}{\multirow{-6}{*}{\textbf{Transfer}}} &   Ours &  23.3M &                &  \textbf{88.4} &  \textbf{68.3} &    \underline{45.1}    &  \textbf{55.6} &  \textbf{71.6} &    \underline{88.6}    &    \underline{61.5} &    \underline{88.9}    &    \underline{63.8}    &    \underline{66.5} &  \textbf{69.8(+0.9)} \\ \hline \hline
\multicolumn{1}{c|}{}                           & Continual-FT                 &                                & 25.5                                  & 81.5                                  & 59.1                                  & 53.2                                  & 64.7                                  & 51.8                                  & 63.2                                  & 64.3                               & 69.7                                  & 31.8                                  & 49.7                               & 55.9                                        \\
\multicolumn{1}{c|}{}                           & LwF                          & 211M                           & 36.3                                  & 86.9                                  & 72.0                                  & 59.0                                  & 73.7                                  & 60.0                                  & 73.6                                  & 74.8                               & 80.0                                  & 37.3                                  & 58.1                               & 64.7                                        \\
\multicolumn{1}{c|}{}                           & iCaRL                        & 211M                           & 35.5                                  & 89.2                                  & 72.2                                  & 60.6                                  & 68.8                                  & 70.0                                  & 78.2                                  & 62.3                               & 81.8                                  & 41.2                                  & 62.5                               & 65.7                                        \\
\multicolumn{1}{c|}{}                           & ZSCL                         & 211M                           & 45.1                                  &   \underline{92.0}                            & 80.1                                  & 64.3                                  &   \underline{79.5}                            & 81.6                                  & \underline{89.6}                                  & \textbf{75.2}                      & 88.9                                  & 64.7                                  & \textbf{68.0}                      & 75.4                                        \\
\multicolumn{1}{c|}{}                           & MoE-Adapter                  & 59.8M                          &   \underline{50.2}                            & 91.9                                  &   \underline{83.1}                            &   \underline{69.4}                            & 78.9                                  &   \underline{84.0}                            &   89.1                   & 73.7                               &   \underline{89.3}                            &   \underline{67.7}                            & 66.9                               &   \underline{76.7}                                  \\
\rowcolor{gray!30} \multicolumn{1}{c|}{\multirow{-6}{*}{\textbf{Average}}}  &  Ours &  23.3M &  \textbf{55.3} &  \textbf{95.6} &  \textbf{84.1} &  \textbf{69.8} &  \textbf{82.3} &  \textbf{85.2} &  \textbf{90.3} &    \underline{75.1} &  \textbf{90.2} & \textbf{67.9} &    \underline{67.8} &  \textbf{78.5(+1.8)} \\  \hline \hline
\multicolumn{1}{c|}{}                           & Continual-FT                 &                                & 31.0                                  & 89.3                                  & 65.8                                  & 67.3                                  & 88.9                                  & 71.1                                  & 85.6                                  & \textbf{99.6}                      & 92.9                                  & 77.3                                  &   \underline{81.1}                         & 77.3                                        \\
\multicolumn{1}{c|}{}                           & LwF                          & 211M                           & 26.3                                  & 87.5                                  & 71.9                                  & 66.6                                  &   79.9                            & 66.9                                  & 83.8                                  & \textbf{99.6}                      & 92.1                                  & 66.1                                  & 80.4                               & 74.6                                        \\
\multicolumn{1}{c|}{}                           & iCaRL                        & 211M                           & 35.8                                  & \underline{93.0}                                  & 77.0                                  & 70.2                                  & 83.3                                  & 88.5                                  & 90.4                                  & 86.7                               & 93.2                                  & 81.2                                  & \textbf{81.9}                      & 80.1                                        \\
\multicolumn{1}{c|}{}                           & ZSCL                         & 211M                           & 40.6                                  &   92.2                            & 81.3                                  & 70.5                                  & 94.8                                  & 90.5                                  &   \underline{91.9}                            & 98.7                               & \textbf{93.9}                         &   \underline{85.3}                            & 80.2                               & 83.6                                        \\
\multicolumn{1}{c|}{}                           & MoE-Adapter                  & 59.8M                          &   \underline{49.8}                            &   92.2                            &   \underline{86.1}                            &   \underline{78.1}                            &   \underline{95.7}                            &   \underline{94.3}                            & 89.5                                  & 98.1                               & 89.9                                  & 81.6                                  & 80.0                               &   \underline{85.0}                                  \\    
\rowcolor{gray!30}\multicolumn{1}{c|}{\multirow{-6}{*}{\textbf{Last}}}     &  Ours &  23.3M  &  \textbf{54.9} &  \textbf{96.4} &  \textbf{87.6} &  \textbf{79.0} &  \textbf{97.6} &  \textbf{96.5} &  \textbf{92.2} &    \underline{98.8} &    \underline{93.6}    &  \textbf{86.6} &  80.7       &  \textbf{87.6(+2.6)}     \\ \hline
\end{tabular}
\caption{On the MTIL benchmark, comparison with state-of-the-art methods in ``Transfer", ``Average", and ``Last" metrics (\%). ``\# Param." is the number of trainable parameters. We highlight the best and second-best methods in \textbf{bold} and \underline{underline} styles.}
\label{MTIL}
\end{table*}

During inference, we input both the input image $x$ and all task descriptions $\mathcal{S} =\{s_i\}_{i=1}^t$ into the MLLMs. The MLLMs then predict the most relevant task identifier including unseen task ($ID_{task}=-1$) by jointly leveraging its visual understanding for $x$ and multimodal alignment with $\mathcal{S}$.
\begin{equation}  
ID_{task}=MLLM(\mathcal{S}, x)
\end{equation}

With semantic understanding rather than statistical information, STR improves robustness to feature space overlaps, noise, and perturbations. Meanwhile, STR only requires to store a compact text description (less than 200 tokens) for each task and does not need training or maintenance of additional model parameters.

\subsection{Fine-tuning Module Dynamic Fusion}To preserve CLIP’s zero-shot transfer ability during continual learning, prior methods\cite{yu2024boosting} processe unseen samples only with pre-trained model, considering that the fine-tuning will impair this ability for original model. Whereas, we observe that there is a consistent enhancement about zero-shot prediction through continual fine tuning, as shown in Figure \ref{zero-shot}, proofing the value of task information. Thus, We dynamically fuse the results of the pre-trained module and the TRGE adapter based on the training phase.

Based on the task identifier, if $x$ belongs to learned tasks, TRGE adapter is directly add with the frozen pre-trained sublayers. If the task is recognized to unseen, we dynamically adjust the weight for adapter with growth factor $\alpha$ and count of learned tasks $t$.  
\begin{equation}  
 y_{out}=\begin{cases}y_{pre}+y_m  & \text{ if } ID_{task} \ne -1  \\y_{pre}+(t \times \alpha) y_m  & \text{ if } ID_{task} = -1\end{cases}
\end{equation}
 where $ y_{pre} $ donates the output of the frozen pre-trained model. As there is no group learned unseen tasks, the inter-group router use Top-k strategy to select groups for them. We suppress fine-tuned weights initially when adapter’s zero-shot transfer ability is weak then increase the weight progressively as training strengthens it's capability.

\begin{table*}[t]
\centering
\small
\setlength{\tabcolsep}{5pt} 
\begin{tabular}{cl|c|ccccccccc|c}
\hline

\multicolumn{2}{c|}{\textbf{Method}}                                       & \textbf{\# Param.}  & $\mathbf{D_1}$ &  $\mathbf{D_4}$ & $\mathbf{D_5}$ & $\mathbf{D_6}$ & $\mathbf{D_7}$ & $\mathbf{D_8}$ & $\mathbf{D_9}$ & $\mathbf{D_{10}}$ & $\mathbf{D_{11}}$ & \textbf{Mean}    \\ \hline

\multicolumn{1}{c|}{}                           & Zero-shot                    &                                                     & 24.3                                  & 44.6                                  & 54.9                                  & 71.0                                  & 88.5                                  & 59.4                                  & 89.0                                  & 64.7                                  & 65.2                               & 62.4                                  \\
\multicolumn{1}{c|}{\multirow{-2}{*}{CLIP}}     & Full Fine-tune               &                                                     & 62.0                                  & 79.5                                  & 98.9                                  & 97.5                                  & 92.7                                  & 99.6                                  & 94.7                                  & 89.6                                  & 81.8                               & 88.5                                  \\ \hline 
\multicolumn{1}{c|}{Qwen-VL}                    & In-Context                   &                                                     & 45.2                                  & 66.8                                  & 86.2                                  & 95.7                                  & 84.7                                  & 78.0                                  & 79.2                                  & 85.6                                  & 65.6                               & 76.3                                  \\ \hline \hline
\multicolumn{1}{c|}{}                           & Continual-FT                 &                                                     &                                       & 32.4                                  & 41.0                                  & 33.2                                  & 57.5                                  & 43.9                                  & 60.7                                  & 20.7                                  & 46.6                               & 42.0                                  \\
\multicolumn{1}{c|}{}                           & LwF                          & 211M                                                &                                       & 39.4                                  & \textbf{56.5}                         & 50.8                                  & 72.6                                  &   \underline{60.4}                            & 75.0                                  & 30.5                                  & 55.9                               & 55.1                                  \\
\multicolumn{1}{c|}{}                           & iCaRL                        & 211M                                                &                                       & 33.0                                  & 44.7                                  & 44.8                                  & 67.8                                  & 45.8                                  & 77.3                                  & 32.1                                  & 60.5                               & 50.7                                  \\
\multicolumn{1}{c|}{}                           & ZSCL                         & 211M                                                &                                       & \textbf{45.7}                         &   \underline{55.8}                            & 67.3                                  & 87.4                                  & \textbf{61.6}                         & 86.7                                  & 60.3                                  & \textbf{66.8}                      & 66.4                                  \\
\multicolumn{1}{c|}{}                           & MoE-Adapter                  & 59.8M                                               &                                       &   \underline{44.7}                         & 55.3                                  &   \underline{68.9}                            &   \underline{88.5}                            & 59.5                                  & \textbf{89.0}                         &   \underline{64.7}                            & 65.5                               &   \underline{67.0}                            \\  \rowcolor{gray!30}\multicolumn{1}{c|}{\multirow{-6}{*}{\textbf{Transfer}}} &  Ours &  23.28M   \textbf{(-61.7\%)}                      &                &    \underline{44.7} &  55.3          &  \textbf{71.2} &  \textbf{88.6} &  60.3          &    \underline{88.6}    &  \textbf{64.8} &    \underline{65.6} &  \textbf{67.4}   \textbf{(+0.4)} \\ \hline \hline
\multicolumn{1}{c|}{}                           & Continual-FT                 &                                                     & 29.1                                  & 59.2                                  & 72.5                                  & 55.5                                  & 62.2                                  & 58.9                                  & 68.7                                  & 33.6                                  & 49.8                               & 54.4                                  \\
\multicolumn{1}{c|}{}                           & LwF                          & 211M                                                & 39.9                                  & 65.0                                  & 81.5                                  & 63.7                                  & 72.6                                  & 69.4                                  & 79.0                                  & 39.1                                  & 58.2                               & 63.1                                  \\
\multicolumn{1}{c|}{}                           & iCaRL                        & 211M                                                & 39.1                                  & 66.6                                  & 76.6                                  & 73.7                                  & 77.2                                  & 56.9                                  & 80.8                                  & 43.0                                  & 62.6                               & 64.0                                  \\
\multicolumn{1}{c|}{}                           & ZSCL                         & 211M                                                & 48.7                                  & \underline{70.3}                                  & \textbf{87.3}                         & 85.3                                  &   \underline{88.6}                            &   \underline{69.8}                            & 87.9                                  & 66.5                                  & \textbf{68.1}                      & 74.7                                  \\
\multicolumn{1}{c|}{}                           & MoE-Adapter                  & 59.8M                                               &   \underline{53.8}                            &   \textbf{75.4}                            &   \underline{86.7}                            &   \underline{87.7}                            & 88.1                                  & 68.3                                  &   \underline{88.3}                            &   \underline{69.5}                            & 67.0                               &   \underline{76.1}                            \\  
\rowcolor{gray!30}\multicolumn{1}{c|}{\multirow{-6}{*}{\textbf{Average}}}  &  Ours &  23.28M   \textbf{(-61.7\%)}                      &  \textbf{55.2} &  \textbf{75.4} &  85.4          &  \textbf{87.9} &  \textbf{90.4} &  \textbf{74.3} &  \textbf{90.2} &  \textbf{69.6} &    \underline{67.0} &  \textbf{77.3}   \textbf{(+1.2)} \\ \hline \hline
\multicolumn{1}{c|}{}                           & Continual-FT                 &                                                     & 35.0                                  & 68.5                                  & 88.9                                  & 73.9                                  & 83.9                                  &   \underline{81.4}                            & 89.8                                  & 82.0                                  &   \underline{79.8}                         & 75.9                                  \\
\multicolumn{1}{c|}{}                           & LwF                          & 211M                                                & 30.3                                  & 67.8                                  & 79.9                                  & 69.7                                  & 82.1                                  &   \underline{81.4}                            & 89.0                                  & 70.8                                  & 79.1                               & 72.2                                  \\
\multicolumn{1}{c|}{}                           & iCaRL                        & 211M                                                & 39.8                                  & 71.4                                  & 83.3                                  & 91.3                                  & 88.7                                  & 68.5                                  & 90.1                                  & 85.9                                  & \textbf{80.6}                      & 77.7                                  \\
\multicolumn{1}{c|}{}                           & ZSCL                         & 211M                                                & 44.6                                  & 71.7                                  & 94.8                                  & 93.3                                  &   \underline{90.2}                            & 80.5                                  &   \underline{90.8}                            & \textbf{90.0}                         & 78.9                               & 81.6                                  \\
\multicolumn{1}{c|}{}                           & MoE-Adapter                  & 59.8M                                               &   \underline{53.8}                            &   \underline{79.3}                            &   \underline{95.7}                            & \textbf{97.1}                         & 87.8                                  & 79.9                                  & 86.8                                  & 86.3                                  & 78.7                               &   \underline{82.8}                            \\ 
\rowcolor{gray!30}\multicolumn{1}{c|}{\multirow{-6}{*}{\textbf{Last}}}     &  Ours & \multicolumn{1}{l|}{23.28M   \textbf{(-61.7\%)}} &  \textbf{55.3} &  \textbf{79.4} &  \textbf{96.2} &    \underline{96.2}    &  \textbf{91.9} &  \textbf{91.8} &  \textbf{93.3} &    \underline{86.4}    &  78.5       &  \textbf{85.5}   \textbf{(+2.7)} \\ \hline
\end{tabular}
\caption{On the MCIL benchmark, comparison with state-of-the-art methods in ``Transfer", ``Average", and ``Last" metrics (\%). ``\# Param." is the number of trainable parameters. We highlight the best and second-best methods in \textbf{bold} and   \underline{underline} styles.}
\label{MCIL}
\end{table*}

\section{Experiment}
\subsection{Experimental Setting}
\subsubsection{Dataset.}We conduct evaluation on MTIL and MCIL benchmark settings. Following\cite{zheng2023preventing}, we conducted experiments across the following 11 datasets denoted as $D_1$ to $D_{11}$: Aircraft\cite{maji2013fine}, Caltech101\cite{fei2004learning}, CIFAR100\cite{krizhevsky2009learning}, DTD\cite{cimpoi2014describing}, EuroSAT\cite{helber2019eurosat}, Flowers\cite{nilsback2008automated}, Food\cite{bossard2014food}, MNIST\cite{deng2012mnist}, OxfordPet\cite{parkhi2012cats}, StanfordCars\cite{krause20133d}, and SUN397\cite{xiao2010sun}. Thses tasks covered multiple classes with significant domain shifts. To prevent potential overlap of class sets between datasets, we simplified the coarse-grained datasets Caltech101 and CIFAR100 as\cite{yu2024select} and conducted experiments on the remaining 9 datasets on MCIL.

\subsubsection{Metrics.} We utilize Transfer, Average and Last to evaluate our method. The Transfer quantifies zero-shot transfer capability on unseen tasks, while the Last metric evaluates the  memorization ability on learned tasks. The Average metric measures mean performance across Transfer and Last.

\subsubsection{Implementation Details.}Following \cite{yu2024boosting}, we employ CLIP ViT-B/16\cite{dosovitskiy2020image} as the backbone with LoRA\cite{hu2022lora} to be expert network. We follow ZSCL's\cite{zheng2023preventing} Order I sequence (Order II is available in the \textbf{Appendix B}). Optimization uses AdamW\cite{loshchilov2017decoupled} with label smoothing\cite{muller2019does}. Unless specified otherwise, expert number $N_e=3$, $k=2$ in Top-k, relevance threshold $\theta=0.3$, and growth factor $\alpha=0.025$. Task identity recognition is handled by Qwen-VL\cite{Qwen-VL}. Each task is trained for 1k iterations on NVIDIA 4090 GPUs.

\begin{table}[]
\centering
\small

\begin{tabular}{cccc|cc}
\hline
Grouping     & Inter-router        & \multicolumn{1}{c}{Last}              &   \multicolumn{1}{c|}{$\Delta$}   & \multicolumn{1}{c}{Average}           & \multicolumn{1}{c}{$\Delta$}     \\ \hline
\ding{55}                         & \ding{55}                         & 85.0                                  &      & 76.7                                  &      \\
\ding{52}                       & \ding{55}                          & 86.8                                  &   +1.8 & 78.0                                  &    +1.3 \\
 \ding{52}   &  \ding{52}   &  \textbf{87.6} &     \textbf{+2.6} &  \textbf{78.5} &    \textbf{+1.8} \\ \hline
\end{tabular}
\caption{Ablation results of each component in TRGE.}
\label{ablation1}
\end{table}

\begin{table}[]
\centering
\small
\setlength{\tabcolsep}{5pt} 
\begin{tabular}{cccc|cc}
\hline
TRGE     & Dynamic Fusion & \multicolumn{1}{c}{Transfer}              &   \multicolumn{1}{c|}{$\Delta$}   & \multicolumn{1}{c}{Average}           & \multicolumn{1}{c}{$\Delta$}     \\ \hline
\ding{55}                         & \ding{55}                         & 69.5                                  &      & 76.7                                  &      \\
\ding{52}                         & \ding{55}                          & 69.5                                  & - & 78.3                                  &    +1.6 \\
 \ding{52}   &  \ding{52}   &  \textbf{69.8} &     \textbf{+0.3} &  \textbf{78.5} &    \textbf{+1.8} \\ \hline
\end{tabular}
\caption{Ablation results of TRGE and dynamic fusion strategy.}
\label{ablation2}
\end{table}

\subsection{Comparison with Methods}
The methods of comparison in our experiments are Continual-FT, LwF\cite{li2017learning}, iCaRL\cite{rebuffi2017icarl}, ZSCL\cite{zheng2023preventing}, MoE-Adapter\cite{yu2024boosting} and the Qwen-VL under the in-context learning.We also present outcomes from applying CLIP to each task via both zero-shot inference and full parameter fine-tuning.  

\subsubsection{Multi-Domain Task-Incremental Learning.}Experimental results on MTIL benchmark are presented in Table \ref{MTIL}. Our method (``Ours") outperforms baselines across all three metrics, achieving average improvements of 1.35\%, 2.35\%, and 3.10\% respectively which demonstrating effectiveness in mitigating catastrophic forgetting and forward forgetting. Notably, our approach achieves superior performance to Qwen-VL using in-context learning with much fewer trainable parameters. It proves that even though MLLMs can achieve the accurate task recognition, they still limited in fine-grained downstream image classification.

\subsubsection{Multi-Domain Class-Incremental Learning.}Under the more challenging MCIL settings, the results are shown in Table \ref{MCIL}. Our method still achieves best or second-best results on most tasks and outperforms other methods in each metric with improvements of 0.57\%, 1.58\% and 3.20\%. It is particularly effective in alleviating catastrophic forgetting, as it enables expert groups to save task knowledge independently. This proves that our method maintains excellent performance even when faced with all learned classes and can correctly recognize the task identifies.

\subsubsection{Computational Cost.}The third column in Table \ref{MTIL} and \ref{MCIL} shows the number of trainable parameters for each method. Our approach achieves superior performance with 61.07\% fewer parameters than MoE-Adapter. This is due to the fact that we only train a small number of experts within a group, rather than training all experts.

\subsection{Ablation Study}
In this section, we evaluate the individual contributions of key components of the TRGE and STR. We analyze the STR on MCIL and others on MTIL.

\subsubsection{Analysis of Two-Level Routing Grouped MoE.}We first analyze the impact of expert grouping and inter-group router on mitigating catastrophic forgetting. As shown in Table \ref{ablation1}, the expert grouping can significantly enhance model's ability against catastrophic forgetting. This is due to grouped experts which independently store knowledge for each task, avoid the interference by subsequent tasks training. Meanwhile, inter-group router effectively achieves tasks collaboration through weighted combination of relevant expert groups which further improves the model's classification accuracy.

\subsubsection{Analysis of Fine-tuning Module Dynamic Fusion.}We further analyze the effectiveness of the dynamic fusion strategy for mitigating forward forgetting. As the results shown in Table \ref{ablation2}, the Transfer metric improved by 0.3. It proves that the task knowledge is valuable for zero-shot prediction and this strategy can collaboratively utilize pre-trained and task knowledge for unseen tasks. The Transfer in second row doesn't improve because the model only with TRGE still relies on the pre-trained CLIP to process unseen samples.

\begin{figure}[t]
  \centering
  \begin{subfigure}[t]{0.49\linewidth}   
    \centering
    \includegraphics[width=\linewidth]{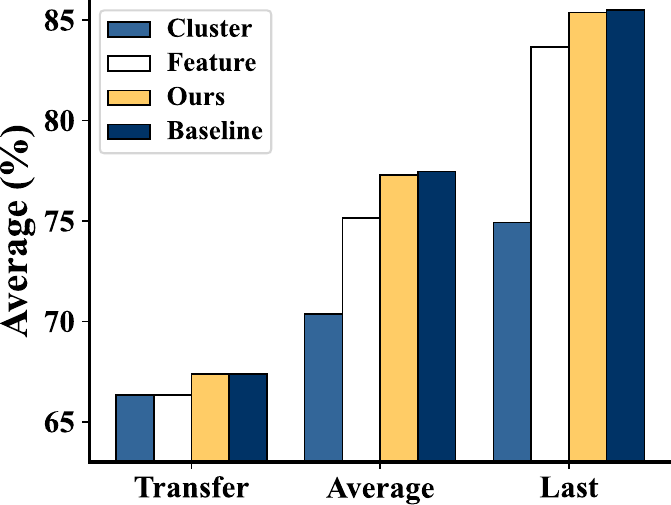}
    \caption{Analysis of STR.}
    \label{SDTR}
  \end{subfigure}
  \begin{subfigure}[t]{0.49\linewidth}   
    \centering
    \includegraphics[width=\linewidth]{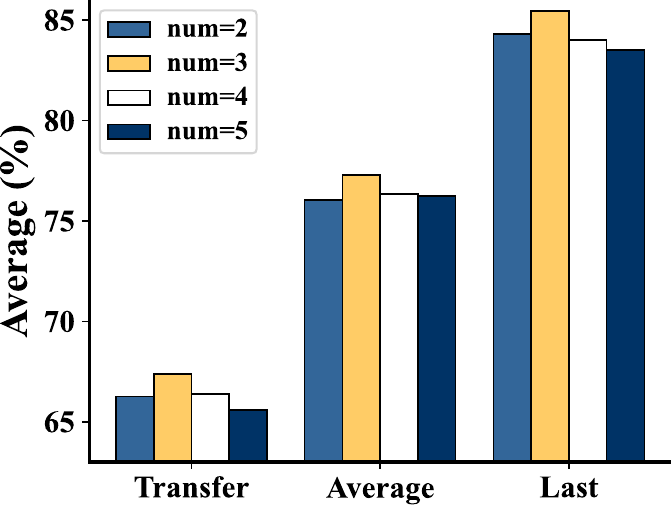}
    \caption{Analysis of Expert Number.}
    \label{Expert_num}
  \end{subfigure}
  \caption{Ablation study results. (a) Model performance with different task identification strategies. (b) Performance variation with different numbers of experts per group.}
  \label{ablation_analysis}
\end{figure}

\subsubsection{Analysis of Semantics-based Task Recognition.}We validated the effectiveness of STR by changing the model's task recognition strategy. We present the outcome from model with given task identifiers as the baseline and compare STR with methods based on clustering\cite{huai2025cl} and feature reconstruction\cite{yu2024boosting}. The results shown in Table \ref{ablation_analysis}(a). STR achieves superior results on all three metrics, proving its high recognition accuracy both in seen and unseen tasks. Notably, the Transfer metric almost matched the baseline performance, indicating that STR achieves nearly 100\% accuracy for unseen tasks. It has been demonstrated that semantic-based methods is more effective than feature-based approaches.

\subsubsection{Analysis of Expert Number.}Figure \ref{ablation_analysis}(b) shows the model's outcomes with different numbers of experts $N_e$ within each group. It can be observed that when $N_e=3$, the model achieves best performance. That's because too little experts limit the model's capacity for knowledge and reduce the specialization of experts. Conversely, too much experts will lead to redundancy and homogenization among experts, increasing risk of overfitting in the routing. The result also validates the rationality of our method to fixing the number of experts per group through expert grouping.

\subsection{Impacts of Hyperparameters}
To further verify the effectiveness of our method, we conducted a parameter sensitivity analysis on MCIL to evaluate the impact of parameters $\theta$ and $\alpha$.

\begin{figure}[t]
  \centering
  \begin{subfigure}[t]{0.49\linewidth}   
    \centering
    \includegraphics[width=\linewidth]{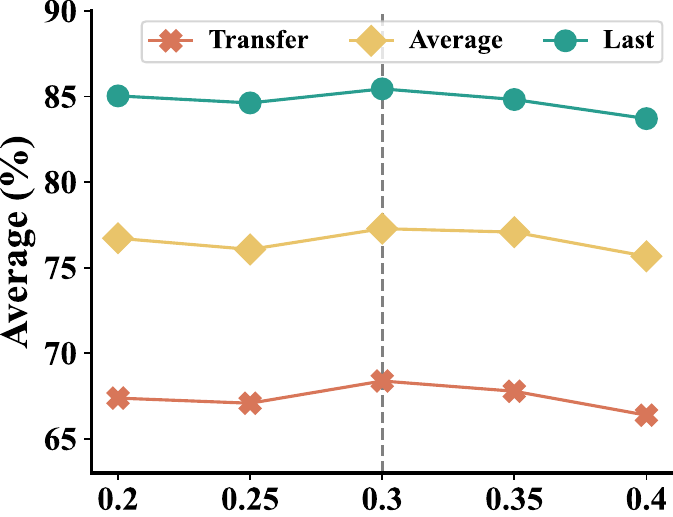}
    \caption{Selection Threshold $\theta$.}
    \label{theta}
  \end{subfigure}
  \begin{subfigure}[t]{0.49\linewidth}   
    \centering
    \includegraphics[width=\linewidth]{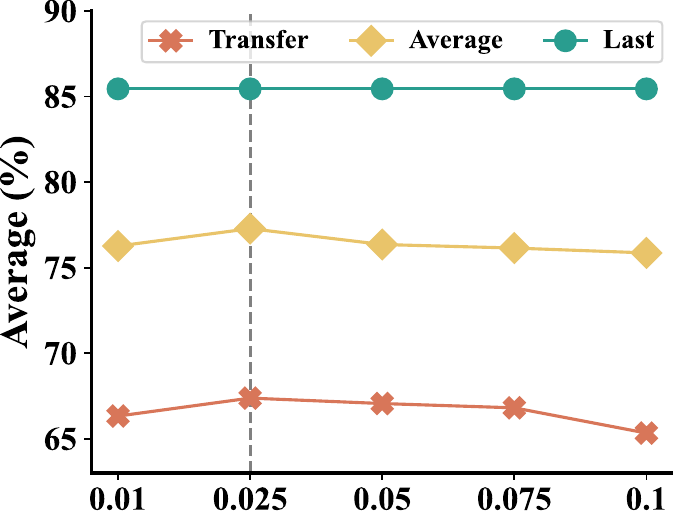}
    \caption{Weight Growth Factor $\alpha$.}
    \label{alpha}
  \end{subfigure}
  \caption{Sensitivity analysis for hyperparameters $\theta$ and $\alpha$.}
  \label{hyper}
\end{figure}

\begin{figure}[t]
\centering
\includegraphics[width=1\columnwidth]{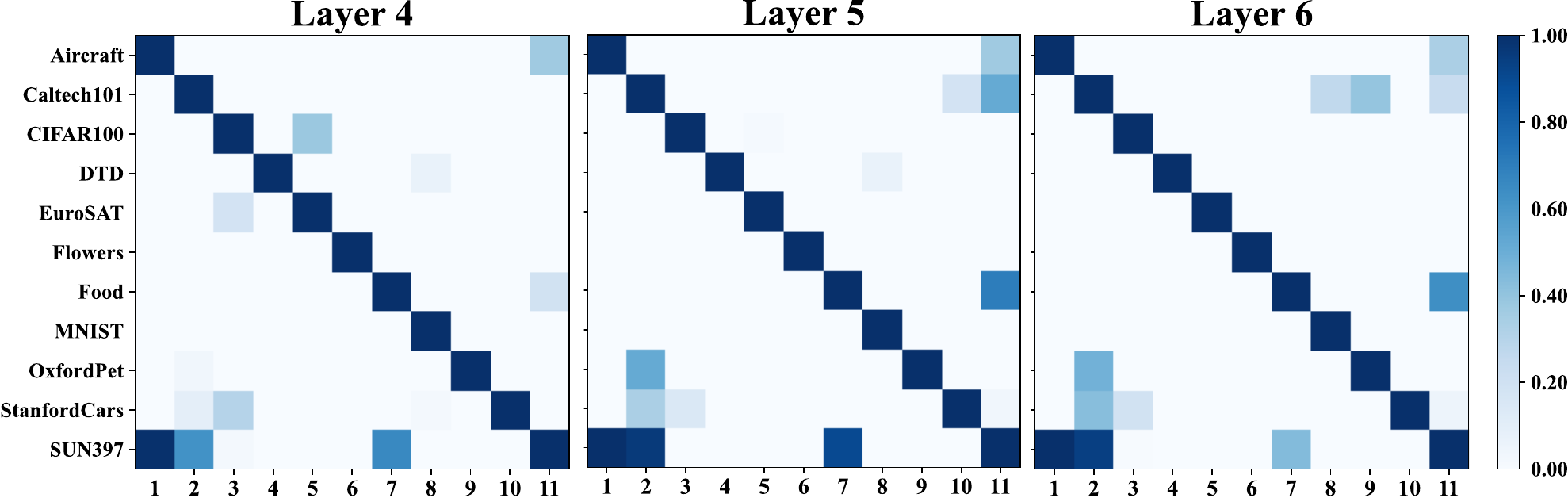}
\caption{Partial visualization of expert selection frequencies. The y-axis represents incremental tasks and the x-axis represents the expert groups.}
\label{visual}
\end{figure}

\subsubsection{Impacts of $\theta$.}Figure \ref{hyper}(a) shows the model performance varying with $\theta$, reaching the optimal performance when $\theta = 0.3$. $\theta$ controls the selection of assistant groups related to input. When $\theta$ is too high, the expert routing will exclude expert groups that are actually related, causing the degradation to the case of no task collaboration. When $\theta$ is too low, irrelevant expert groups will be activated, introducing noise interference and leading to a decline in model performance. 

\subsubsection{Impacts of $\alpha$.}Figure \ref{hyper}(b) shows the performance varying with $\alpha$, reaching the optimal point at $\alpha=0.025$. The $\alpha$ controls the growth rate for the weights of TRGE adapter in zero-shot prediction. A too small $\alpha$ will suppress the contribution of TRGE adapter, failing to effectively utilize task knowledge. While a too large $\alpha$ allows the fine-tuned model to dominate before it has sufficiently learned, thereby harming the model performance. $\alpha=0.025$ can reasonably boost the fine-tuned weights during training, achieving a favorable integration of task knowledge and pre-trained knowledge.

\subsection{Visualization Analysis}
Figure \ref{visual} shows a visualization of the groups selection frequency in partial layers, and the results for all layers can be found in the \textbf{Appendix D}. The y-axis represents the test dataset, and the x-axis represents the index of expert group. It can be seen that the selection frequency of expert groups exhibits a certain symmetry. This proves that inter-layer routing can select relevant assistant groups for tasks, effectively achieving cross-task knowledge transfer.

\section{Conclusion}
To address the dual challenges of catastrophic forgetting and forward forgetting in multi-domain continual learning, we propose a Two-Level Routing Grouped Mixture-of-Experts (TRGE) method. TRGE incrementally extends an expert group on the frozen CLIP encoder for each new task, and dynamically weights the output of relevant expert groups based on the distance between input and task prototypes for inter-group routing, achieving tasks' collaboration. To improve the accuracy of task recognition, we innovatively propose a semantics-based task recognition method. Additionally, for unseen samples, the output of adapter and pre-trained module are dynamically weighted combined to enhance zero-shot transfer capability. Extensive experiments across different settings demonstrate that our method achieves superior performance in both anti-forgetting and zero-shot generalization with fewer parameters.

\bibliography{aaai2026}
\end{document}